\documentclass{article}

\PassOptionsToPackage{numbers, compress}{natbib}

\usepackage[preprint]{neurips_2025}

\usepackage[utf8]{inputenc} 
\usepackage[T1]{fontenc}    
\usepackage{hyperref}       
\usepackage{url}            
\usepackage{booktabs}       
\usepackage{amsfonts}       
\usepackage{nicefrac}       
\usepackage{microtype}      
\usepackage{xcolor}         
\usepackage{amsthm}
\usepackage{amsmath}
\usepackage{graphicx}
\usepackage{caption}
\usepackage{subfigure}
\usepackage{subfloat}
\usepackage{float}
\usepackage{algorithm}
\usepackage{algorithmic}
\usepackage{stfloats}
\usepackage[capitalize,noabbrev]{cleveref}
\usepackage{enumitem}
\usepackage{comment}
\usepackage{dsfont}
\usepackage{bbm}
\usepackage{wrapfig}
\usepackage[most]{tcolorbox}
\usepackage{enumitem}
\usepackage{xcolor}
\usepackage{lipsum} 

\usepackage{amsmath,amsfonts,bm}

\newcommand{\captiona}{{\em (a)}}
\newcommand{\captionb}{{\em (b)}}
\newcommand{\captionc}{{\em (c)}}

\def\eqref#1{equation~\ref{#1}}

\def\1{\bm{1}}

\DeclareMathAlphabet{\mathsfit}{\encodingdefault}{\sfdefault}{m}{sl}
\SetMathAlphabet{\mathsfit}{bold}{\encodingdefault}{\sfdefault}{bx}{n}

\newcommand{\E}{\mathbb{E}}

\newcommand{\R}{\mathbb{R}}

\DeclareMathOperator{\Tr}{Tr}

\newcommand{\bu}{\boldsymbol{u}}

\newcommand{\bx}{\boldsymbol{x}}

\newcommand{\bA}{\boldsymbol{A}}
\newcommand{\bB}{\boldsymbol{B}}
\newcommand{\bC}{\boldsymbol{C}}
\newcommand{\bD}{\boldsymbol{D}}

\newcommand{\bH}{\boldsymbol{H}}

\newcommand{\bS}{\boldsymbol{S}}

\newcommand{\btheta}{\boldsymbol{\theta}}

\newcommand{\cD}{\mathcal{D}}

\newcommand{\cL}{\mathcal{L}}

\newcommand{\norm}[1]{\left\lVert#1\right\rVert}

\definecolor{myPurple}{HTML}{6E3C76}
\definecolor{myPurpleLight}{HTML}{F5EDF8}
\definecolor{middlegrey}{rgb}{0.75,0.75,0.75}
\definecolor{lightblue}{HTML}{F9FEFE}

\title{New Evidence of the Two-Phase Learning Dynamics of Neural Networks}

\author{
Zhanpeng Zhou\textsuperscript{1}$^{\ast}$, Yongyi Yang\textsuperscript{2}, Mahito Sugiyama\textsuperscript{3,4}, Junchi Yan\textsuperscript{1}\thanks{Corresponding authors. This work extends our workshop paper: \emph{On the Cone Effect in the Learning Dynamics} (accepted by ICLR 2025 Workshop DeLTa). \normalsize}\\
\textsuperscript{1}Shanghai Jiao Tong University, \textsuperscript{2}University of Michigan, 
\textsuperscript{3}National Institute of Informatics, \\
\textsuperscript{4}The Graduate University for Advanced Studies, SOKENDAI\\
  \texttt{\{zzp1012,yanjunchi\}@sjtu.edu.cn}
}

\begin{document}

\maketitle

\begin{abstract}
Understanding how deep neural networks learn remains a fundamental challenge in modern machine learning. A growing body of evidence suggests that training dynamics undergo a distinct phase transition, yet our understanding of this transition is still incomplete. In this paper, we introduce an interval-wise perspective that compares network states across a time window, revealing two new phenomena that illuminate the two-phase nature of deep learning. i) \textbf{The Chaos Effect.} By injecting an imperceptibly small parameter perturbation at various stages, we show that the response of the network to the perturbation exhibits a transition from chaotic to stable, suggesting there is an early critical period where the network is highly sensitive to initial conditions; ii) \textbf{The Cone Effect.} Tracking the evolution of the empirical Neural Tangent Kernel (eNTK), we find that after this transition point the model's functional trajectory is confined to a narrow cone-shaped subset: while the kernel continues to change, it gets trapped into a tight angular region. Together, these effects provide a structural, dynamical view of how deep networks transition from sensitive exploration to stable refinement during training.
\end{abstract}

\section{Introduction}\label{sec:introduction}

Modern neural networks have become widely used in a broad range of fields~\citep{dlapplication-medical,dlapplication-robots,dlapplication-weather}. 
However, a thorough understanding of how their capabilities and behaviors are developed throughout the training process remains incomplete, especially for deep models~\citep{learning-dynamics-survey,concept-learning-dynamics}. 
Many recent studies have suggested, either implicitly or explicitly, that there is a \emph{phase transition} point during the NN training, where the model's properties and behaviors undergo substantial shifts before and after this time point. 
For example, \citet{edge-of-stability,damian2023selfstabilization,eos-analysis} showed that during training, the network first enters a progressive sharpening phase, and after which, the sharpness stabilizes and remains roughly constant for the rest of the training.
\citet{critical-period} identified a \emph{critical learning period} early in training, during which exposure to low-quality data can cause irreversible damage, while similar exposure later in training can be reversed.

Despite abundant evidence for the \emph{two-phase phenomenon}~\citep{concept-learning-dynamics,edge-of-stability,eos-analysis,lmc,llfc,critical-period}, a complete characterization and understanding of this phenomenon still lags behind. 
Moreover, most existing studies adopt a ``point-wise'' perspective: they primarily focus on examining specific properties of the network at isolated time points. 
This perspective, while informative, offers only a static snapshot of the model's behavior, and does not capture the temporal dynamics of learning: how a property emerges, evolves, or vanishes as training progresses. 

\begin{figure}[tb!]
    \centering
    \includegraphics[width=0.85\linewidth]{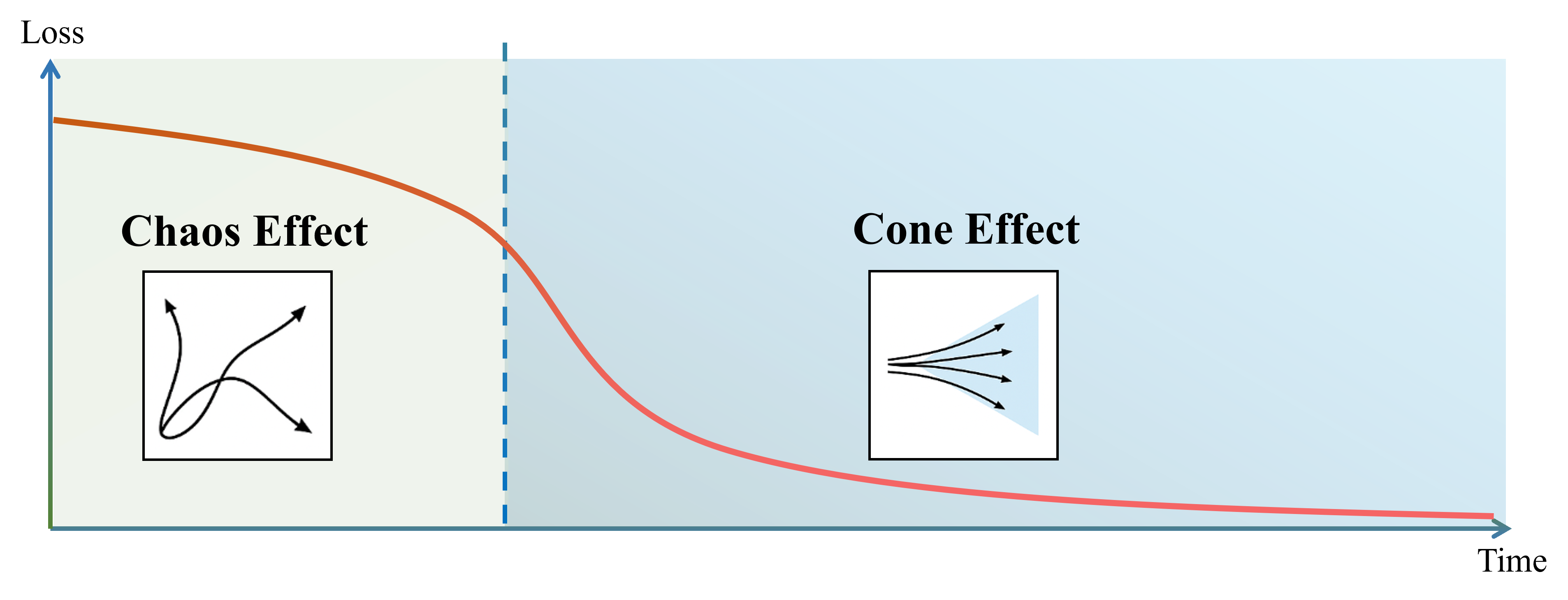}
    \caption{An illustration of the two-phase training dynamics. The optimization trajectory initially passes through a chaotic training phase, termed as the \emph{chaos effect}; then undergoes a more stable, non-chaotic regime, namely the \emph{cone effect}.}
    \label{fig:intro}
\end{figure}

In this paper, to gain a deeper understanding of the two-phase phenomenon, we introduce two new empirical observations that exhibit characteristics of the phase transition. 
Crucially, these are what we call ``interval-wise'' phenomena: rather than analyzing a property at specific time points, we compare the model's behavior across two different time points of training. 
We show that this novel approach reveals patterns that are otherwise invisible to point-wise analysis and offers new insights into the learning dynamics of neural networks. 
Specifically, we identify and investigate two distinct behaviors: \emph{the Chaos Effect} and \emph{the Cone Effect}. See \Cref{fig:intro} for an illustration.

\textbullet~\textbf{The Chaos Effect.} 
First, we observe that the learning dynamics of neural networks transition from a chaotic to a stable, non-chaotic regime during training.
Specifically, we train two networks that are initialized identically and trained with the same stochastic gradient noise.
At a specific time $t_0$, we apply a small perturbation to the parameters of one of them, and then we compare the resulting parameters at a later time $t_1$.
Particularly, we observe an \emph{inflection point} during the training process.
We find that when $t_0$ is in the early stage of training, specifically before the inflection point, even a tiny perturbation leads to a significant divergence from the original training trajectory.
This phenomenon indicates a high sensitivity of learned parameters to initial conditions, which is a hallmark of chaotic systems in physics.
However, if $t_0$ is later in the training (after the inflection point), the divergence is minimal, suggesting that the system becomes increasingly stable as training progresses.


\textbullet~\textbf{The Cone Effect.} Second, we discover that after the early training phase, the learning dynamics of neural networks keep constrained in a narrow cone in the function space.
Specifically, we train a network, and starting from a chosen time point $\tau$, we track the empirical Neural Tangent Kernel (eNTK) at later time steps and measure their deviation from the eNTK at $\tau$. 
We observe that when $\tau$ is sufficiently large, the subsequent eNTKs remain confined within a narrow cone around the eNTKs at time $\tau$. 
In contrast, if the $\tau$ is chosen in the early stages of training, the eNTKs experience chaotic and large unstructured changes over time, and no such confinement is observed. 

\textbf{Our contributions.} 
In summary, we present an empirical study of neural network learning dynamics, with the aim of guiding future theoretical and experimental investigations in deep learning. 
Our main contributions are:
\begin{itemize}[topsep=0in,leftmargin=*]
    \item \emph{Interval-Wise Analysis Framework.} We propose a novel interval-wise framework for analyzing learning dynamics in neural networks;
    \item \emph{Chaos and Cone Effects.} We identify and characterize two distinct phenomena: the Chaos Effect and the Cone Effect. Both exhibit clear two-phase behavior;
    \item \emph{Structural Insights into Training Evolution.} We demonstrate how these effects uncover new structural properties of the neural network’s temporal evolution.
\end{itemize}

\textbf{Roadmap.}
This rest of the paper is organized as follows.
In \cref{sec:related-work}, we recall the existing related works. 
In \cref{sec:preliminaries}, we introduce our notation and the quantitative measures used in our empirical study.
In \cref{sec:chaos}, we present and analyze the Chaos Effect. 
In \cref{sec:cone}, we describe the Cone Effect.
Finally, in \cref{sec:conclusion}, we summarize our findings and discuss potential limitations.



\section{Related Work}\label{sec:related-work}

As discussed in \Cref{sec:introduction}, many studies have suggested neural network training undergoes a phase transition in practice. 
In our view, this body of work can be roughly grouped into the following four categories:

\textbullet~\textbf{Transition of the First-order Quantities.} 
A recent line of research investigated phase transitions in neural network training using gradient-based metrics.
For example, \citet{deep-vs-kernel} identified a clear branching point in the evolution velocity of the empirical Neural Tangent Kernel (eNTK), transitioning from a fast to a slow regime.
\citet{break-even-point} observed that the largest eigenvalue of the gradient covariance matrix increases monotonically in the early phase and then enters a regime of sustained oscillation.
    
\textbullet~\textbf{Transition of the Second-order Quantities} 
Extensive works analyzed the Hessian spectrum of neural networks to characterize training phase transitions. 
For example, \citet{edge-of-stability,eos-analysis,damian2023selfstabilization,Frankle2020The} reported a two-phase pattern in the largest Hessian eigenvalue (the ``sharpness'') during gradient descent: an initial phase of steady growth in sharpness accompanied by smooth loss reduction, followed by a plateau where sharpness fluctuates around a critical threshold.  
\citet{hessian-density} studied the full Hessian spectrum, clearly revealing a structural phase transition. 
\citet{gradient-descent-tiny-space} showed that the subspace spanned by the top Hessian eigenvectors stabilizes shortly after initialization, indicating a transition in the dominant eigenspace.

\textbullet~\textbf{Transition of the Training Trajectories.} 
Another line of research focused on the phase transition behavior from the perspective of optimization trajectory and loss landscape geometry.
For instance, \citet{lmc,Frankle2020The,deep-vs-kernel,llfc} introduced a \emph{spawning method}: a network is initialized and trained for a few epochs, then spawned into two copies that continue training independently under different sources of SGD randomness (e.g., mini-batch order and data augmentation). 
They showed that when spawning occurs late in training, the two models, despite far in Euclidean distance, remain in the same basin of the loss landscape; whereas early spawning yields models in distinct basins. 
\citet{singh2025the} analyzed the directionality of the optimization path and identified a corresponding phase transition from this directional viewpoint.

\textbullet~\textbf{Transition in Behaviors/Ability.} From a more functional perspective, \cite{concept-learning-dynamics} demonstrate that neural networks learn concepts sequentially based on their signal strength. There exists a critical moment when one concept has been fully learned while another remains unlearned, clearly marking a phase boundary. \cite{critical-period} analyze how models respond to temporary corruptions in training data. They identify a ``critical period'' early in training during which such temporary corruption can cause permanent damage, while similar interventions later in training are reversible.



\section{Preliminary and Methodology}\label{sec:preliminaries}

\paragraph{Basic Notations.}
Throughout this paper, we focus on a classification task.  Denote $[k]= \{1, 2, \cdots, k\}$.
Let $\cD=\{(\bx_i,y_i)\}_{i=1}^n$ be the training set of size $n$, where $\bx_i \in \R^{d_0}$ represents the $i$-th input and $y_i\in [c]$ represents the corresponding target.
Here, $c$ is the number of classes.
Let $f: \cD \times \R^p \to \R$ be the NN model, and thus $f(\bx, \btheta) \in \R$ denotes the output of model $f$ on the input $\bx$ with parameter $\btheta \in \R^p$.
Let $\ell(f(\bx_i, \btheta), y_i)$ be the loss at the $i$-th data point, simplified to $\ell_i(\btheta)$.
The total loss over the dataset $\cD$ is then denoted as $\cL_{\cD}(\btheta) = \frac{1}{n}\sum_{i=1}^n \ell_i(\btheta)$.
We also use $\operatorname{Err}_{\mathcal D} (\boldsymbol{\theta})$/$\operatorname{Acc}_{\mathcal D} (\boldsymbol{\theta})$ to denote the classification error/accuracy of the network $f(\boldsymbol{\theta}; \cdot)$ on the training set $\cD$. 

Additionally, we use bold lowercase letters (e.g., $\bx$) to denote vectors, and bold uppercase letters (e.g., $\bA$) to represent matrices. 
For a matrix $\bA$, let $\norm{\bA}_2$, $\norm{\bA}_F$, and $\Tr({\boldsymbol{A}})$ denote its operator norm, Frobenius norm, and trace respectively.

\paragraph{Parameter Dissimilarity.}
Following \citet{singh2025the}, we first introduce the \emph{parameter dissimilarity} to measure the directionality of the optimization process. Specifically, given the training trajectory consisting of a sequence of checkpoints $\{\btheta_t\}_{t=0}^T$, we use the pairwise cosine dissimilarity to capture the directional aspect of the trajectory.
For any two time points $i, j \in [T]^2$, we define
\begin{align}
    (\bC)_{i,j} := 1 - \cos \langle \text{vec}(\btheta_i), \text{vec}(\btheta_j) \rangle = 1 - \langle \text{vec}(\btheta_i), \text{vec}(\btheta_j) \rangle / (\norm{\text{vec}(\btheta_j)}_2 \norm{\text{vec}(\btheta_j)}_2),
\end{align}
where $\text{vec}(\btheta)$ denotes the flattened parameters of the network.
We note that the sequence of checkpoints $\{\btheta_t\}_{t=1}^T$ represents only a subset of the entire training trajectory that encountered in practice, sampled at intervals of $k$ points.
The pairwise parameter dissimilarity matrix $\bC$ will serve as a qualitative measure for analyzing the directionality of the training dynamics.

\paragraph{Kernel Distance.}
Similar to \citet{deep-vs-kernel}, we use the \emph{kernel distance} to quantify the evolution of neural networks in the function space. 
Consider minimizing the total loss $\cL_{\cD}(\btheta)$ using gradient flow; the evolution of the network function $f_t(\bx_i) =f(\bx_i, \btheta_t)$ can be written as:
\begin{equation}
    \frac{\mathrm{d}  f_t(\bx_i)}{\mathrm{d}  t} = -\sum_{j=1}^n \left\langle \frac{\partial f_t(\bx_i)}{\partial \btheta}, \frac{\partial f_t(\bx_j)}{\partial \btheta} \right\rangle \frac{\partial \cL_{\cD}(\btheta_t)}{\partial f_t(\bx_i)}. \label{eq:func_learn_single}
\end{equation}
Denoting $\bu_t=\{f_t(\boldsymbol{x}_i)\}_{i=1}^n \in \R^n$ as the network outputs for all inputs, then a more compact form of \cref{eq:func_learn_single} is given by:
\begin{equation}
\label{eq: eNTK}
\frac{\mathrm{d} \bu_t}{\mathrm{d}  t}= - \bH(\btheta_t) \nabla_{\bu_t}\cL(\btheta_t), \quad\text{where } \left(\bH(\btheta_t)\right)_{i,j} =   \displaystyle \left\langle \frac{\partial f_t(\bx_i)}{\partial \btheta}, \frac{\partial f_t(\bx_j)}{\partial \btheta} \right\rangle,
\end{equation}
Here, the kernel matrix $\boldsymbol{H}(\btheta_t) \in \R^{n\times n}$ is often termed as the \emph{empirical neural tangent kernel} (eNTK).
To study the evolution of neural network in function space, we measure the pairwise distance between the eNTK matrices at two different time points, namely \emph{kernel distance}.
For two time points $i, j \in [T]^2$, we define:
\begin{equation}\label{eq: kernel distance}
    (\bS)_{i, j} := 1 - \frac{\langle\bH(\btheta_i), \bH(\btheta_j)\rangle}{\norm{\bH(\btheta)_i}_F \norm{\bH(\btheta_j)}_F}.
\end{equation}
We will use the kernel distance matrix $\bS$ to analyze the training dynamics in function space. 

\paragraph{Loss Barriers.}
In addition to the directional and functional aspect, we also investigate the geometry of the neural network's loss landscape, focusing specifically on the regions encountered during the training process. 
In particular, we examine the \emph{loss barriers}~\citep{lmc,ainsworth2023git} between any two points along the training trajectory.
For any two points $i, j \in [T]^2$, we define:
\begin{align}
    (\bB)_{i, j} := \max_{\alpha} \cL_{\cD'} (\alpha \btheta_i + (1-\alpha) \btheta_j) - \frac{1}{2} (\cL_{\cD'}(\btheta_i) + \cL_{\cD'}(\btheta_j)),
\end{align}
where $\cD'$ denotes the unseen test set.
Indeed, the loss barrier is typically high for two independently trained neural networks, indicating that the two models reside in different and isolated ``valleys''.
Once the loss barrier approaches zero, i.e., $(\bB)_{i, j} \approx 0$, we say the two models $\btheta_i$ and $\btheta_j$ are linearly connected in the loss landscape~\citep{lmc,llfc}.

\paragraph{Disagreement Rate.}
Lastly, we concern about the similarity of the outputs at two points along the optimization trajectory.
Specifically, we introduce the \emph{disagreement rate} on the test data for any two points $i, j \in [T]^2$: 
\begin{align}
    (\bD)_{i,j} := \E_{\bx \in \cD'} [\1 (f(\bx, \btheta_i) \neq f(\bx, \btheta_j))],
\end{align}
where $\1 (\cdot)$ is the indicator function and $\cD'$ is the test set.
Notably, \citet{jiang2022assessing} demonstrated that the test error of deep models can be approximated by the disagreement rate between two independently trained models on the same dataset, i.e., $\operatorname{Err}_{\cD'}(\btheta) \approx \operatorname{Err}_{\cD'}(\btheta') \approx \E_{\bx \in \cD'} [\1 (f(\bx, \btheta) \neq f(\bx, \btheta'))]$.


\paragraph{Main Experimental Setup.}
We train the VGG-16 architecture~\citep{simonyan2015vgg} and the ResNet-20 architecture~\citep{kaiming2016residual} on the CIFAR-10 dataset. 
Data augmentation techniques include \texttt{random horizontal flips} and \texttt{random $32\times32$ pixel crops}. 
Optimization is done using SGD with momentum (momentum set to 0.9). 
A weight decay of $1\times 10^{-4}$ is applied. 
The learning rate is initialized at $0.1$ and is dropped by $10$ times at $80$ and $120$ epochs. The total number of epochs is $160$. 


\section{The Chaos Effect: Sensitivity of Learning Dynamics to Small Perturbations}
\label{sec:chaos}

Our investigation of the two-phase learning dynamics is motivated by a simple question:
\begin{center}
    \emph{How do injected noise influence the learning dynamics of neural networks? Is the training trajectory robust to small perturbation encountered during optimization.}
\end{center}

\begin{wrapfigure}[19]{R}{0.36\textwidth}
    \vspace{-5pt}
    \centering
    \includegraphics[width=0.95\linewidth]{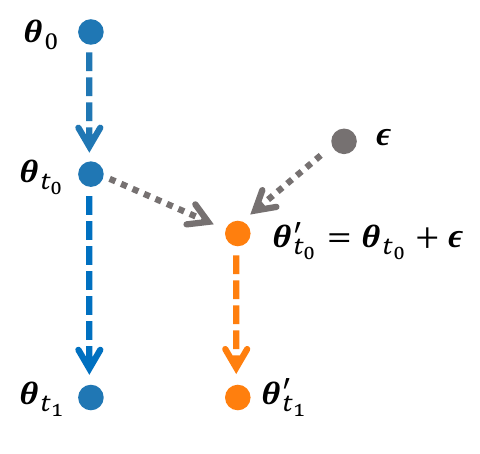}
    \caption{
    \textbf{The illustration of the injected perturbation.}
    $\btheta_0$ denotes the initialization.
    For both $\btheta$ and $\btheta'$, the same stochastic gradient noise are applied during training.
    }
    \label{fig:perturbation_method}
    \vspace{-5pt}
\end{wrapfigure}
To answer this question, in the section, we empirically investigate the sensitivity of neural network learning dynamics to small perturbations. 
Surprisingly, we find that an inflection point emerges during the training process.
We observe that even a tiny perturbation applied before the inflection point can cause a significant divergence from the original optimization trajectory. 
In contrast, applying perturbations after the inflection point have far less impact.
Thus we conjecture that the inflection point severs as a hallmark of the transition from a chaotic to an non-chaotic regime.
We refer to this phenomenon as the \emph{chaos effect}.


\paragraph{Experimental Design.}
We train two networks with identical initializations and the same stochastic gradient noise. 
However, at a specific time $t_0$, we introduce a small perturbation $\boldsymbol\epsilon$ to the parameters of one network, such that $\btheta_{t_0}' = \btheta_{t_0} + \boldsymbol\epsilon$.
We then compare the resulting models at a later time $t_1$ ( $t_1 > t_0$), with the parameters $\btheta_{t_1}'$ and $\btheta_{t_1}$ respectively.
The experimental design is illustrated in \cref{fig:perturbation_method}.
We consider a tiny perturbation here, where $\norm{\boldsymbol\epsilon}_0 = 10^{-7}$.
We vary the time $t_0, t_1$ and compare the two resulting models using different metrics. 
The results are presented below.

\begin{figure}[tb!]
    \centering
    \includegraphics[width=0.99\linewidth]{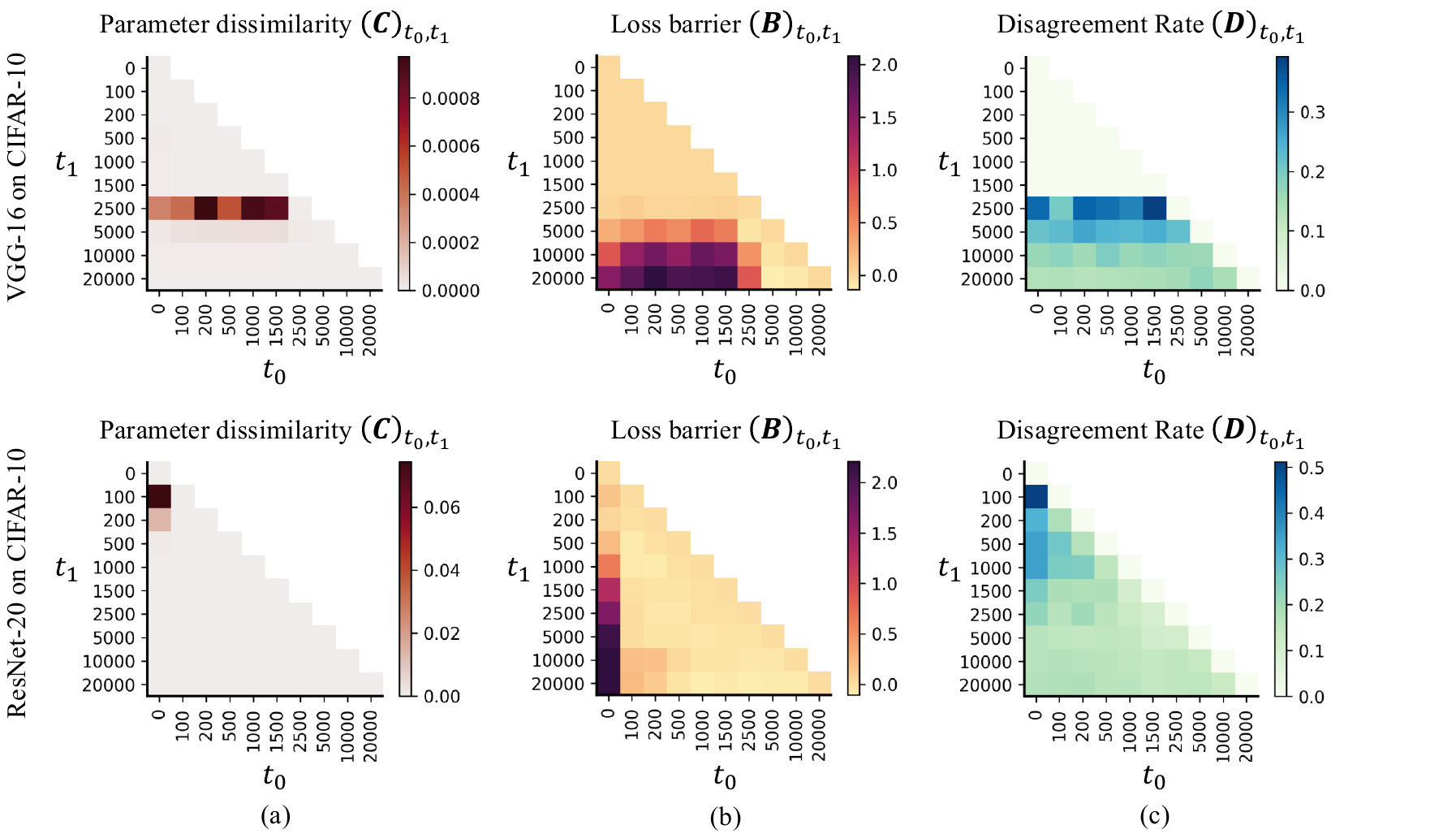}
    \caption{\textbf{The sensitivity of learning dynamics to tiny perturbations.}
    We apply the perturbation $\boldsymbol\epsilon$ at the time point $t_0$ and compare resulting models at $t_1$ with different metrics.
    We set $\norm{\boldsymbol{\epsilon}}_0 = 10^{-7}$.
    Our results are reported for VGG-16 and ResNet-20 on CIFAR-10.
    \textbf{(a)} 
    The parameter dissimilarity $(\bC)_{t_0, t_1}$.
    \textbf{(b)}
    The loss barrier $(\bB)_{t_0, t_1}$.
    \textbf{(c)}
    The disagreement rate $(\bD)_{t_0, t_1}$.
    Note that the $t_0$ and $t_1$ are presented in iterations, not epochs.
    }
    \label{fig:chaos}
\end{figure}

\paragraph{Finding I. Optimization trajectory changes its direction at an inflection point.}
As shown in \cref{fig:chaos}~\captiona, we first present the parameter dissimilarity for any pair of $t_0$ and $t_1$ (with $t_1 \geq t_0$).
Notably, across different training settings, there exists a specific time $t_1$ at which the value of $(\bC)_{t_0, t_1}$ remains relatively high for all choices of $t_0$. 
For example, in the case of VGG-16 on CIFAR-10, the dissimilarity $(\bC)_{t_0, t_1}$ consistently reaches its maximum when $t_1 = 2500$ iteration, regardless of the value of $t_0$.
Typically, a high value of $(\bC)_{t_0, t_1}$ indicates the directional change along the optimization trajectory.
Therefore, it is evident that the optimization trajectory changes its direction at a fixed point, namely the \emph{inflection point}.

\paragraph{Finding II. Tiny perturbations applied before the inflection point leads to significant loss barriers and disagreement rate.}
In \cref{fig:chaos}~\captionb, we also report the loss barrier between each pair of time points $t_0, t_1$. 
We observe that even a tiny perturbation ($\norm{\boldsymbol{\epsilon}}_{0} = 10^{-7}$) applied at a early time point $t_0$ could result in a substantial loss barrier between the resulting parameters $\btheta_{t_1}$ and $\btheta_{t_1}'$ in the later stage of training.
This indicates that the two solutions likely reside in different, isolated ``valleys'' of the loss landscape.
In \cref{fig:chaos}~\captionc, we further evaluate the disagreement rate between $\btheta_{t_1}$ and $\btheta_{t_1}'$.
The observed high disagreement rate firmly validates the functional dissimilarity between $\btheta_{t_1}$ and $\btheta_{t_1}'$.
Together, these results indicate that the learning dynamics pass through a chaotic regime during the early phase of training, where small perturbations might lead to pronounced loss barriers and functional divergence later on, namely the \emph{chaos effect}.

\paragraph{Conjecture I. The inflection points marks the transition from a chaotic to an non-chaotic regime.}
Interestingly, the results observed for loss barriers and disagreement rate exhibit patterns similar to those for parameter dissimilarity.
Taking VGG-16 on CIFAR-10 as an example, we observe that a significant loss barrier $(\bB)_{t_0, t_1}$ emerges only when $t_0 \leq 2500$ iterations and $t_1 > 2500$ iterations.
Similar observations are also noted for the disagreement rate.
Recall that the $2500$ iteration marks the inflection point for VGG-16 on CIFAR-10.
Therefore, we conjecture that the inflection point serves as a hallmark of the transition from a chaotic to a non-chaotic training regime.

\begin{figure}[tb!]
    \centering
    \includegraphics[width=0.86\linewidth]{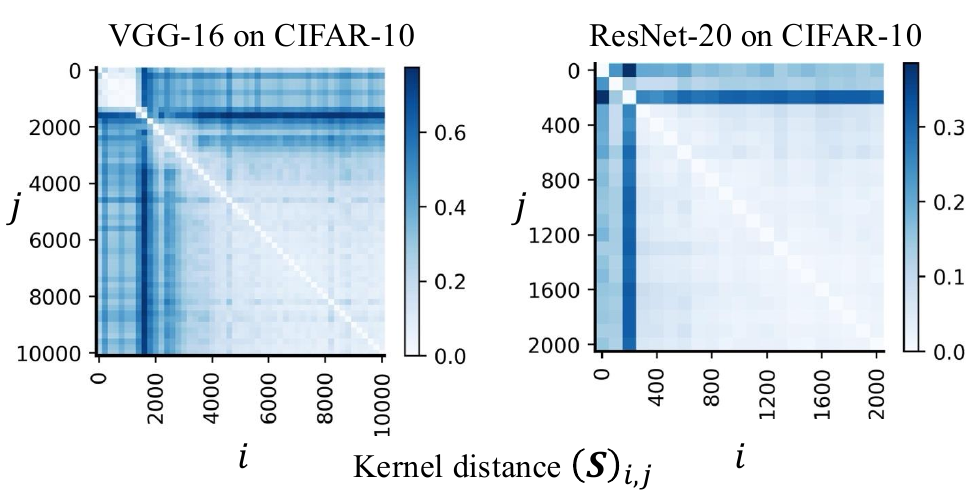}
    \caption{\textbf{The kernel distance between every pair of two points at the optimization trajectory $\{\btheta_t\}_{t=1}^T$.}
    Our results are reported for both VGG-16 and ResNet-20 on CIFAR-10.
    Note that the $i$ and $j$ are presented in iterations, not epochs.
    }
    \label{fig:kernel_single}
    \vspace{-5pt}
\end{figure}

To further verify this conjecture, we compute the kernel distance between every pair of points along the optimization trajectory within a single training run. 
Specifically, we train the neural network and obtain a sequence of checkpoints $\{\btheta_t\}_{t=1}^T$.
Then we measure the kernel distance for any two points $i, j \in [T]^2$, i.e., $(\bS)_{i,j}$.
As shown in \cref{fig:kernel_single}, we observe that the eNTKs evolve significantly during the early phase of training, indicating a chaotic regime. 
Subsequently, the evolution of the eNTKs stabilizes, signifying a transition to a non-chaotic training phase.
Notably, the transition point in the evolution of the eNTKs aligns with the inflection points identified in earlier experiments. 
For example, in the case of ResNet-20 on CIFAR-10, both the eNTK transition and the inflection point occur around $100\sim 500$ iteration.
These results provide strong support for our conjecture that the inflection point marks the boundary between chaotic and non-chaotic training phases.



\vspace{1em}
\begin{tcolorbox}[
    colback=myPurpleLight,
    colframe=myPurple,
    coltitle=white,
    fonttitle=\bfseries,
    title=Section 4 Key Takeaways:,
    boxrule=1pt,
    arc=4pt,
    left=6pt,
    right=6pt,
    top=6pt,
    bottom=6pt,
    enhanced,
    width=\textwidth
]
\begin{itemize}[leftmargin=*]
    \item Inflection points emerge during training, marking significant changes in the direction of the optimization trajectory.
    \item Tiny perturbations applied before the inflection point lead to substantial divergence later in training, indicating a chaotic regime in the early phase.
    \item The inflection point signifies a transition from a chaotic to a stable, non-chaotic training phase.
\end{itemize}
\end{tcolorbox}
\vspace{1em}

\begin{figure}[tb!]
    \centering
    \includegraphics[width=0.99\linewidth]{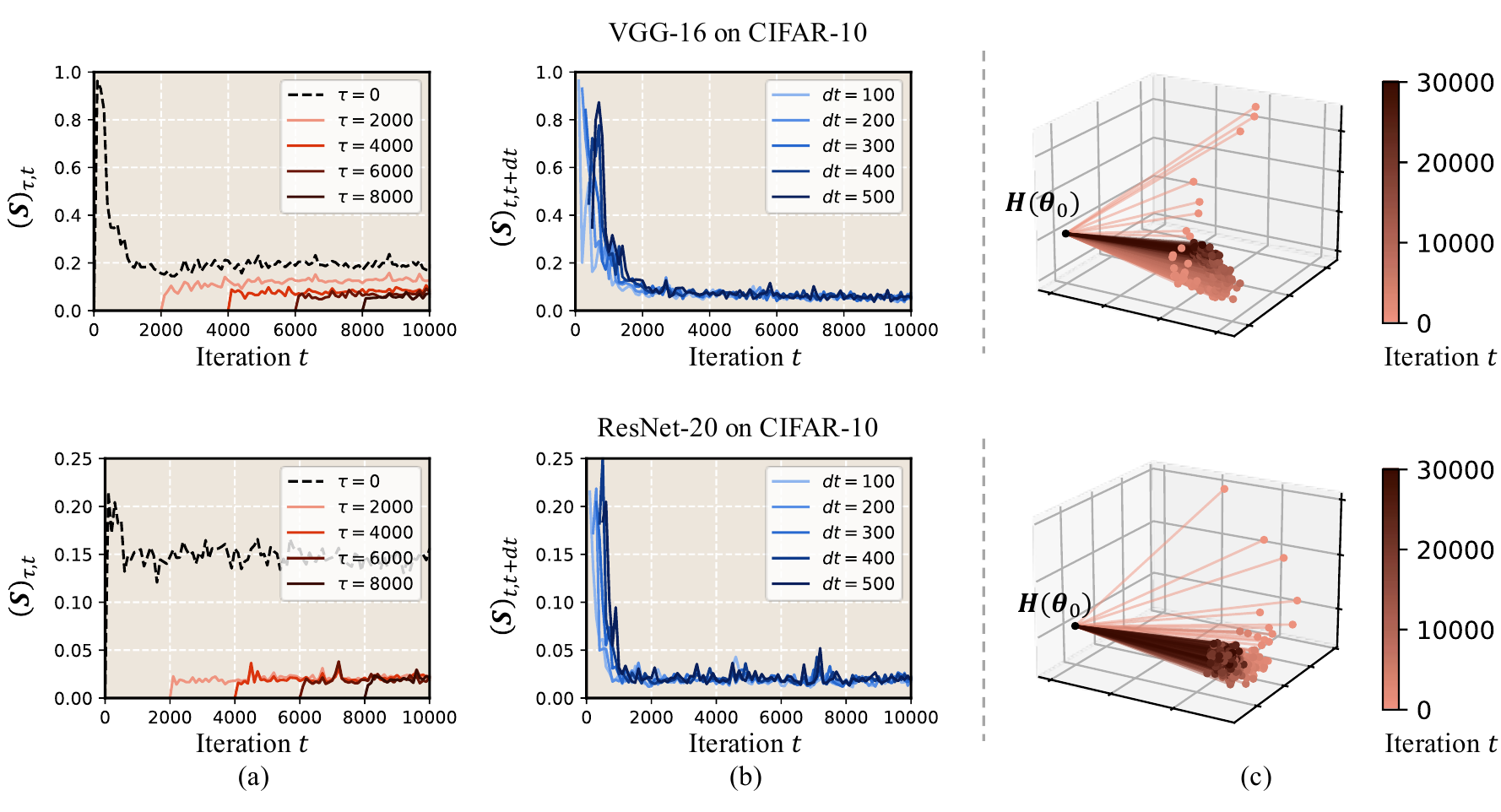}
    \caption{\textbf{Constrained learning dynamics in the second phase.}
    \textbf{(a)} 
    The kernel distance between the current iterate $\btheta_t$ and a reference point $\btheta_{\tau}$ v.s. training iteration $t$, where $\tau$ is varied.
    \textbf{(b)}
    The kernel distance between two adjacent iterates $\btheta_t$ and $\btheta_{t+dt}$ vs. training iteration $t$, where $dt$ is varied.
    \textbf{(c)}
    The visualization of the changes of the eNTK matrices $\bH(\btheta_t)$.
    The black dot represents position of the eNTK matrix at initialization, i.e., $\bH(\btheta_0)$.
    The other dot represents the relative position of $\bH(\btheta_t)$ at $t>0$, with darker color indicating larger iteration.
    }
    \label{fig:cone_effect}
\end{figure}

\section{The Cone Effect: Constrained Learning Dynamics in the Second Phase}
\label{sec:cone}

We have seen that the learning dynamics of neural networks undergo a transition from a highly chaotic to a more stable, non-chaotic phase, marked by a consistent inflection point.
During the chaotic phase, the network evolves rapidly in function space, as evidenced by the significant changes in the eNTKs (see \cref{fig:kernel_single}).
However, a natural question arises: 
\begin{center}
    \emph{What characterizes the training dynamics during the subsequent stable, non-chaotic phase?}
\end{center}

In this section, we dig deeper into the learning dynamics in the ``second'' phase.
Surprisingly, we note that, contrary to the typical assumption of the \emph{lazy training regime}, which we will elaborate on shortly, the neural network continues to evolve. 
However, this evolution is no longer unconstrained; instead, it is confined within a narrow, ``cone''-like region in function space, a phenomenon we refer to as the \emph{cone effect}. 

\paragraph{The Lazy Regime.}
Numerous theoretical studies~\citep{du2018gradient,li2018learning,du2019gradient,allenzhu2019convergence,zou2020gradient} proved that over-parameterized models can achieve zero training loss with minimal parameter variation.
Moreover, \citet{jacot2018NTK,yang2019scaling,arora2019on,lee2019wide} showed that the learning dynamics of infinitely wide neural networks can be captured by a frozen kernel at initialization and are considered linear in theory.
This behavior, often termed as \emph{lazy regime}, typically occurs in over-parameterized models with large initialization and is considered undesirable in practice~\cite{Chizat2019lazytraining}.

Previous studies often hypothesized that the learning dynamics during the later stages of training~\citep{NEURIPS2018_6651526b,deep-vs-kernel,li2022what} or during fine-tuning~\citep{wortsman2022model,Wortsman_2022_CVPR} are approximately linear, suggesting the presence of a \emph{lazy training regime}.
However, recent works~\citep{ortiz-jimenez2023task,DBLP:conf/icml/ZhouC00Y24,edge-of-stability,eos-analysis} have challenged this view, showing that linearized dynamics alone are insufficient to capture the behavior of neural networks in the later phases of training.
Our observations are consistent with these findings: rather than exhibiting lazy dynamics, we identify the \emph{cone effect} during the second training phase. 
This constrained yet non-trivial evolution plays a crucial role in shaping the network's generalization ability.


\textbf{Beyond the Lazy Regime: The Cone Effect.}
Through the kernel distance, we delve into the second phase of the learning dynamics.
First, we compute the distance between the kernel matrices at two adjacent points $\btheta_t$ and $\btheta_{t+dt}$.
In \cref{fig:cone_effect}~\captionb, we observe that, across different values of $dt$, the kernel distance $(\bS)_{t, t+dt}$ between adjacent iterates $\btheta_t$ and $\btheta_{t+dt}$ is significant in the early training phase and then drops quickly to a low but non-negligible value.
This result aligns with our previous results in \cref{fig:chaos}, where in the early training phase, the model evolves significantly in the function space.
However, surprisingly, we note that in the later training phase, the values of $(\bS)_{t, t+dt}$ are upper-bounded by the same value for different $dt$.
One possible explanation of this phenomenon is that during the second training phase, the eNTK matrix evolves in a constrained space.

To validate this, we further measure how the distance between the kernel matrices at the current iterate $\btheta_t$ and a referent point $\btheta_{\tau}$ changes during training.
As shown in \cref{fig:cone_effect}~\captiona, for different referent points $\tau\in \{2000, 4000, 6000, 8000\}$, the kernel distance between the current iterate and the reference point, i.e., $S(\btheta_t, \btheta_{\tau}$, first increases and then keeps nearly constant in training. This result supports our claim and suggests that during later training phase, beyond the lazy regime, the model operates in a constrained function space.
The visualization in \cref{fig:cone_effect}~\captionc~ further confirms the existence of the cone effect, where a clear ``cone'' pattern is observed during the evolution of eNTK matrices.

\begin{figure}[tb!]
    \centering
    \includegraphics[width=0.85\linewidth]{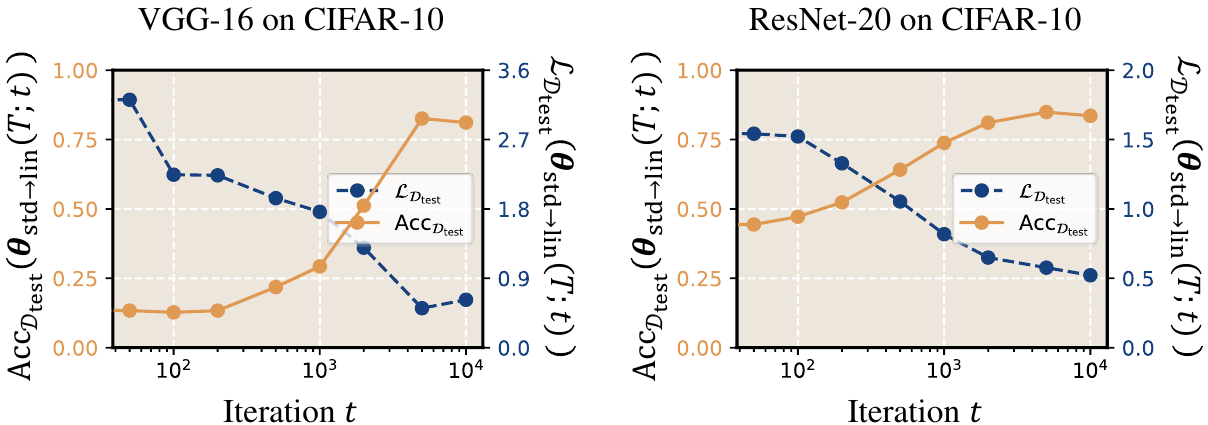}
    \caption{\textbf{The non-linear advantage of the cone effect.}
    Test accuracy $\text{Acc}_{\cD_{\rm test}}(\btheta_{\rm std \to lin}(T; t))$ \textbf{(left)} and Test loss $\cL_{\cD_{\rm test}}(\btheta_{\rm std \to lin}(T; t))$ vs. the switching iteration $t$.
    $\btheta_{\rm std \to lin}(T; t)$ represents the model initially trained with standard method up to iteration $t$, followed by linearized training up to iteration $T$, where $T$ is set to $10^{4}$.
    }
    \label{fig:advantage}
\end{figure}

\paragraph{The Non-linear Advantages of the Cone Effect.}
Despite the model evolving in a constrained function space in the second phase, it still provides significant advantages over completely lazy regime.
To verify this, we consider a ``switching'' training method: we first train a neural network with standardized training method, and then switch to the linearized training (corresponding to the completely lazy regime) until $T$ iterations.
We vary the switching point $t$ and obtain different solutions $\btheta_{\rm std \to lin}(T; t)$.
In \cref{fig:advantage}, we observe that the test performance of $\btheta_{\rm std \to lin}(T; t)$ generally increases with $t$, especially when $t > 2000$. 
This result implies that the cone effect in the later training phase still offers significant advantages over the entirely lazy regime.


\begin{tcolorbox}[
    colback=myPurpleLight,
    colframe=myPurple,
    coltitle=white,
    fonttitle=\bfseries,
    title=Section 5 Key Takeaways:,
    boxrule=1pt,
    arc=4pt,
    left=6pt,
    right=6pt,
    top=6pt,
    bottom=6pt,
    enhanced,
    width=\textwidth
]
\begin{itemize}[leftmargin=*]
    \item Despite the stable, non-chaotic dynamics in the later training phase, the neural networks the neural network continues to evolve within a narrow, constrained ``cone''-like region of the function space, namely the cone effect.
    \item Compared to purely linearized training, the cone effect offers significant advantages for the performance of the final solution.
\end{itemize}
\end{tcolorbox}

\section{Conclusion and Limitations}
\label{sec:conclusion}

In this paper, we introduced an interval-wise perspective on neural network training dynamics, moving beyond traditional point-wise analyses to reveal new insights into the learning process. 
Through this lens, we identified two novel empirical phenomena that characterize a two-phase transition in deep learning: the Chaos Effect and the Cone Effect. 
The Chaos Effect highlights a critical early period during which the training trajectory is highly sensitive to small perturbations, indicating chaotic behavior. 
In contrast, the Cone Effect demonstrates that, after this transition point, the model's functional evolution becomes increasingly constrained within a narrow region of the function space, even though learning continues.

Together, these findings suggest a transition from an exploratory, unstable phase to a more stable, refinement-oriented phase during training. 
Our interval-wise analysis framework not only captures dynamic behaviors missed by point-wise approaches but also opens new directions for understanding and improving the training of deep neural networks. 
We hope this work inspires further research into the temporal structure of learning dynamics and the underlying mechanisms driving phase transitions in modern deep learning.

\textbf{Limitations.}
We note that our current work primarily focuses on empirical findings, despite strongly related to optimization theory and NTK theory, we defer a thorough theoretical analysis to future work.
We also note that our current experiments mainly focus on image classification tasks, though aligning
with existing empirical studies on training dynamics~\citep{deep-vs-kernel}.
We leave the exploration of empirical evidence beyond image classification as future direction.


{
\small
\bibliography{ref}
\bibliographystyle{plainnat}
}




\end{document}